\documentclass{article} 
\usepackage{nips15submit_e,times}
\usepackage{hyperref}
\usepackage{url}
\usepackage{authblk}
\usepackage{amsmath,graphicx,booktabs}
\usepackage{epstopdf,amssymb}
\usepackage{multirow}
\usepackage{fixltx2e}
\usepackage{subfigure}
\usepackage{float}
\usepackage{titlesec}
\usepackage{bm}
\usepackage{mathrsfs}

\title{\textbf{Deformable Registration Using Average Geometric Transformations for Brain MR Images}} 

\author{Yongpei Zhu$^{1}$, Zicong Zhou$^{2}$, Guojun Liao$^{2}$, Kehong Yuan$^{1*}$\\
	$^{1}$Graduate School at Shenzhen, Tsinghua University, Shenzhen 518055, China.\\
	$^{2}$The University of Texas at Arlington, Arlington 76019, USA.\\
	*Corresponding author: Kehong Yuan (e-mail: yuankh@sz.tsinghua.edu.cn)\\
	\texttt{zhuyp17@mails.tsinghua.edu.cn}\\
}

\nipsfinalcopy 

\begin{document}
	
	\maketitle
	
	\begin{abstract}
		Accurate registration of medical images is vital for doctor's diagnosis and quantitative analysis. In this paper, we propose a new deformable medical image registration method based on average geometric transformations and VoxelMorph CNN architecture. We compute the differential geometric information including Jacobian determinant(JD) and the curl vector(CV) of diffeomorphic registration field and use them as multi-channel of VoxelMorph CNN for second train. In addition, we use the average transformation to construct a standard brain MRI atlas which can be used as fixed image. We verify our method on two datasets including ADNI dataset and MRBrainS18 Challenge dataset, and obtain excellent improvement on MR image registration with average Dice scores and non-negative Jacobian locations compared with MIT¡¯s original method. The experimental results show the method can achieve better performance in brain MRI diagnosis.\\
		
	\end{abstract}
	
%
%
%
%


\section{Introduction}
\label{sec1}
Image registration is a common task in MR image analysis and has been an active research topic in many areas, which aligns the image space into a common anatomical space. The deformable registration strategy calculates the dense correspondence between two images and generates the affine transformations for global alignment, which are usually much slower and have higher degrees of freedom.
There are two types of methods in deformable registration, including classical registration methods (non-learning-based) and learning-based methods. Classical registration methods solve the optimization problems on spatial of deformation (the space of displacement vector field), including elastic-type models [1], b-splines [7], statistical parameter mapping [17], Demons [9] or discrete methods [5]. Diffeomorphic transformations have undergone extensive advancement, resulting in state-of-the-art tools such as LDDMM, DARTEL and SyN. But these tools demand amounts of time and computational resources for a given image pair on a CPU.

The emergence of deep learning enables neural networks to penetrate into medical image registration. Recently, some methods are proposed to train neural networks that map a pair of input images to output deformation. Balakrishnan et al. [11] performed a rigorous analysis of their method and presented the results on a complete MR volume. They studied MR brain image registration based on 3D slices and the computational efficiency of this method is also higher than the existing general methods. They proposed an unsupervised learning network called VoxelMorph[15], using a pair of 160x192x224 3d images as input. They tested their method on eight publicly available MRI brain image datasets. They also present a probabilistic generative model and derive an unsupervised learning-based inference algorithm [3][4] to provide diffeomorphic guarantees and uncertainty estimates.

In this paper, we propose a new deformable medical image registration method based on differential geometric information which trained on original network in [4] and get the state-of-the-art performance among else methods before to the best of our knowledge. The constructions of this paper are:
1. we propose a new deformable medical image registration method based on differential geometric information including the Jacobian determinant(JD) and the curl vector(CV) of diffeomorphic registration field on the VoxelMorph CNN architecture in the paper[4].

2.we also build a registration template for 7 MRI image samples according to average of differential geometric transformation.

\section{Method}

\subsection{The Deformation Method}
In deformable registration problem, it is essential to find diffeomorphic registration field $\phi$. Here, we introduce the deformation method that can generate JD and CV by the grid generation (namely registration field $\phi$), which will be used in our later methods. Diffeomorphism is an active research topic in differential geometry [13]. JD and CV play an important role in determining a diffeomorphism. Since JD has the same direct physical meaning in grid generation which represents the size of grid cell and CV represents the grid cell rotations, the deformation method was applied successfully to grid generation and registration problems of MRI brain image[2].

We use MATLAB to generate the grid images (diffeomorphic registration field $\phi$) using the deformation method on the gradient and gray value of original image. And we extract the images formed by JD and CV information from the grid images (Fig.1(c)(d)). The following Fig.1(b)(c)(d) show the grid image, JD and CV image of MRI brain respectively, which present geometric deformation features of the image, especially highlight the change of morphological features of CSF, GM, WM. According to manifold learning [10], CNN can extract the manifold features from images. So, we extract JD and CV images from generated known $\phi$ and use them as input of CNN for better extracting manifold features.

\subsection{Variational Method for Image Registration}
Here, we use variational method to construct diffeomorphisms $\phi$, through controlling directly the Jacobian determinant and the curl vector of a transformation, which shows the important role of the JD and the CV in determining a diffeomorphism. It is called the Variational Method[13]. Let $\phi=({\phi}_{1},{\phi}_{2},{\phi}_{3})$, $u=({u}_{1},{u}_{2},{u}_{3})$, $f=(f_{1},f_{2},f_{3})$,${g}_{0}=(g_{01},g_{02},g_{03})$, we define the similarity measure in 3D as:
\begin{equation}
ssd=\frac{1}{2} \iiint_{\mathrm{\Omega}}[{(J(\phi)-f_0)}^{2}+{(curl(\phi)-\mathbf{g_0})}^{2}]dV
\end{equation}
Here, $f_0$ is the prescribed Jacobian determinant monitor function, and $g_0$ is the prescribed curl vector monitor function. So, the problem we study now is to construct numerically a transformation $\phi:\Omega\to\Omega$, namely, $J({\phi})=f_0,curl(\phi)=\mathbf{g_0}$.

For two different images ${I}_{1}(x,y,z)$ and ${I}_{2}(x,y,z)$, image registration process is to fnd a transformation $\phi$ such that $I_1(\phi)=I_2$. Here, a new registration algorithm based on variational method is developed. Given the moving image T and the fixed image R in 3D, we determine a registration transformation $\phi$ by iteratively minimizing a similarity measure:
\begin{equation}
SSD=\frac{1}{2}{\iiint}_{\varOmega}{(T({\phi}(x,y,z))-R(x,y,z))}^{2}dV
\end{equation}

\section{Proposed Framework}
\subsection{Training Strategy Based on Average JD and CV}
The following Fig.2 shows the proposed framework and the training strategy of our method. Given a same atlas or reference volume, N volumes of our training set $({y}_{1},{y}_{2},...,{y}_{N})$ were used as the input of the network, and output the approximate posterior probability parameters including the velocity field mean ${\mu}_{(z|x,y)}$ and variance ${\varSigma}_{z|x,y}$. The velocity field z is sampled and transformed to a diffeomorphic deformation field ${\phi}_z$ using squaring and scaling layers. Finally, we use a spatial transform function warps $y_i$ to $y_i\circ{\phi}_z$. At registration stage, we generate the image of deformation filed ${\phi}_z$ by evaluating $def_{\phi}(x,y)$ during registration between original each two volumes (x,$y_i$). As for known generated registration filed $\phi$, we extract JD and CV images (note CV images have 3 volumes) in the first training stage and use them as channels of original networks for second train. So, we can generate N pairs of $(JD_i,CV_i1,CV_i2,CV_i3)$, for each moving image subject of train set, and combine them with corresponding JD and CV generated from $\phi_{z_i}$ as 5 channels, and for fixed image subject, we combine it with the average image of $(JD_i,CV_i1,CV_i2,CV_i3)$, namely $(\sum_{\mathrm{i}}{JD_{i}}/N,\sum_{\mathrm{i}}{CV_{i1}}/N,\sum_{\mathrm{i}}{CV_{i2}}/N,\sum_{\mathrm{i}}{CV_{i3}}/N)$ as 5 channels, finally combined moving and fixed images are input into Voxelmorph to train again and then we test registration between pairs of unseen subjects from testing set. The loss can also be the original composed of two terms by estimating the variational lower bound using SGD methods in MIT's paper. To make comparison in following experiment, we also get JD and CV from diffeomorphisms $\phi$ which generated directly using deformation method on the gradient and gray value of input moving volumes described in section 2.1, but not from the registration filed $\phi$ in the first training stage, then perform the same process as above.

The following Fig.3 shows the deformation filed $\phi$ of initial training stage and generated JD and CV image from $\phi$. Fig.3 (a) shows the RGB image of $\phi$ and Fig.3 (b)(c)(d) shows each channel of Fig.3(a). Fig.3 (e) shows the generated JD of $\phi$. It is important to note that the curl of the image has three components(channels) coming from different axes in 3D. Fig.3(f) represents the RGB image of generated
CV and each channel of it is shown in Fig.3(g)(h)(i). In this work, we use generated JD and CV as channels of the training model to train again based on the VoxelMorph CNN architecture and we get better performance than train only once in MIT's paper [21] which we describe in the following section.

\begin{figure}[ht]
\centering
\includegraphics[scale=1,width=0.5\textwidth]{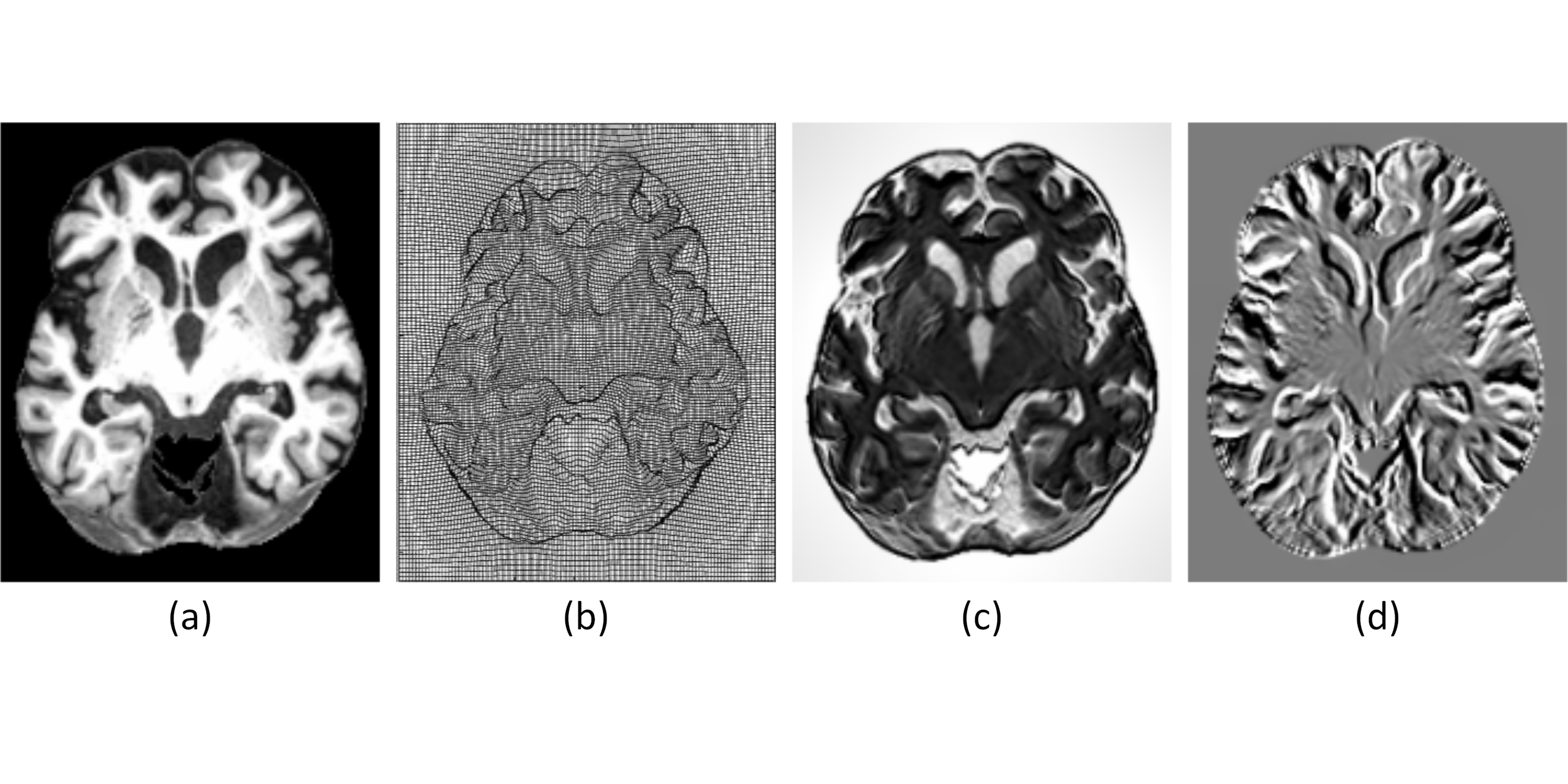}
\caption{Generated images based on JD: (a) The original T1 image, (b) The grid image, (c) The image formed by JD, (d) The image formed by CV.}
\end{figure}

\subsection{Average Transformations}
We use average a set of transformations[16] by averaging their JD and CV. Given a series of transformations ${\phi}_{i}$ for $i=1,2,...,N$, we construct their average in the following steps:

(1)Let $f_0=\frac{1}{N}{{\sum^{\mathrm{N}}_{\mathrm{i=1}}}{\mathrm{J({\phi}_{i})}}}$ and $g_0=\frac{1}{N}{\sum^{\mathrm{N}}_{\mathrm{i=1}}}{\mathrm{curl({\phi}_{i})}}$ as in Eq.(1).

(2)Use Eq. (1) to calculate a transformation $\phi$ such that ${\mathrm{J({\phi})={f}_{0}}}$, and  ${\mathrm{curl({\phi})=g_0}}$.

(3)We define this $\phi$ as the average of ${\phi}_{i}$ for $i=1,2,...,N$.

The average deformation has certain geometric meaning: the local size change ratios modeled by $J({\phi}_{i})$ and the local rotations modeled by $curl({\phi}_{i})$ are averaged to average transformation. So, we apply average transformations to medical image registration. This is also the reason why we combine average of JD and CV with Fixed image as 5 channels in the above framework to predict better registration filed $\phi$. Because it synthesizes all the information of ${\phi}_{i}$ which are obtained in the first training stage and it can improve the accuracy of medical image registration.

\begin{figure}[ht]
\centering
\includegraphics[scale=1,width=0.5\textwidth]{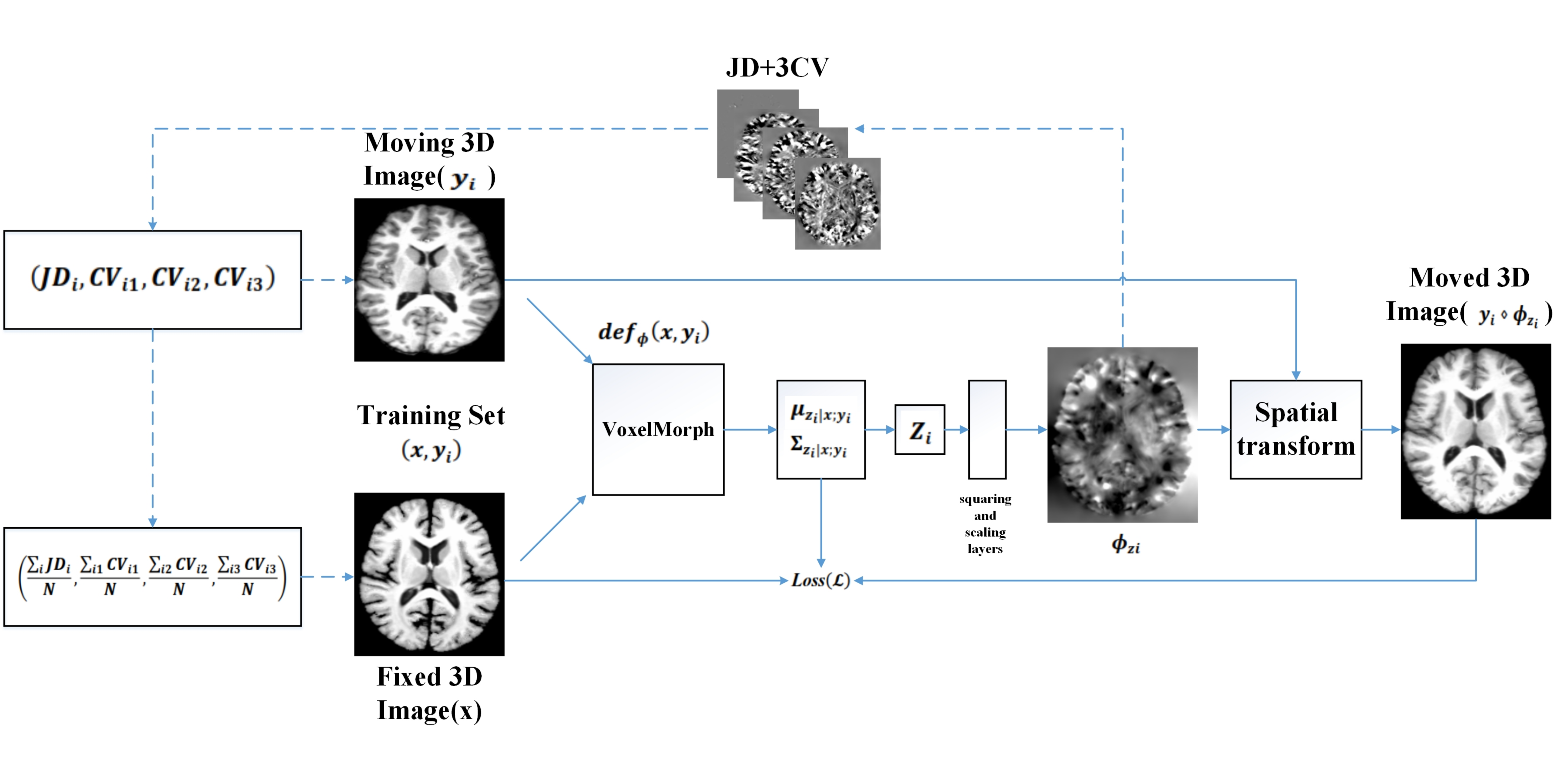}
\caption{The proposed framework of our method.}
\end{figure}

\subsection{Construction of Atlas(Template)}

Here, we want to construct an unbiased template using average of transformations according to Fig.4.

Step1: for our dataset with N subjects, we take one subject $y_i$ out as the initial template, then register  ${y}_{i}$ to all N images ${y}_{j}$ for $j=1,2,...,i,...,N$ to get all deformation filed $\mathrm{\phi}_{ij}$ with Voxelmorph CNN.

Step2: find the average transformation ${\mathrm{avg}}_{i}$=$avg(\mathrm{\phi}_{ij})( j=1,2,...,i,...,N)$ of all the deformation filed by the Average of Transformations method described in above. This step will get N average transformations ${\mathrm{avg}}_{i}$ for $i=1,2,...,N$.

Step3: Warp ${y}_{i}$ on the average transformation ${\mathrm{avg}}_{i}$ to get the temporary templates ${\hat{T}{emplate}_{i}}$=${y}_{i}({\mathrm{avg}}_{i} (\mathrm{\phi}_{ij}))$ for $i=1,2,...,N$. This step will get N temporary templates ${\hat{T}{emplate}_{i}}$.

Step 4: repeat Step-1 to Step-3 on biased temporary templates ${\hat{T}{emplate}_{i}}$ to reduce their bias. So we get a new group of average transformation ${\mathrm{Avg}}_{i}$ and new temporary templates ${\hat{T}_{i}}={\hat{T}{emplate}_{i}}({\mathrm{Avg}}_{i})$.And we will do above steps until the average transformation ${\mathrm{avg}}_{i}$ is close to the identify map Id (unit orthogonal grid which corresponds to unit orthogonal map).

\section{Experiments}
\subsection{Datasets, Preprocessing and Evaluation Criteria}
\subsubsection{Datasets and Preprocessing}
We validate our method on two datasets including ADNI[8] and MRBrainS18.
ADNI We use one dataset from ADNI including 199 T1-weighted brain MRI scans. And we split the dataset into 159,20 and 20(8:1:1) volumes for train, validation, and test sets respectively.

MRBrainS18 We apply the average transformation method on this dataset. Seven brain MRI scans (size:240x240x48) of MRBrainS18 with manual segmentations are provided for training and no data are provided publicly for testing. The data can be downloaded from http://mrbrains18.isi.uu.nl/.

Before the CNN training stage, standard preprocessing steps for structural brain MRI include the following steps, including skull stripping, resampling and affine spatial normalization using FreeSurfer[6]. Segmentation maps including 29 anatomical structures obtained using FreeSurfer for evaluation.

\begin{figure}[ht]
\centering
\includegraphics[scale=1,width=0.5\textwidth]{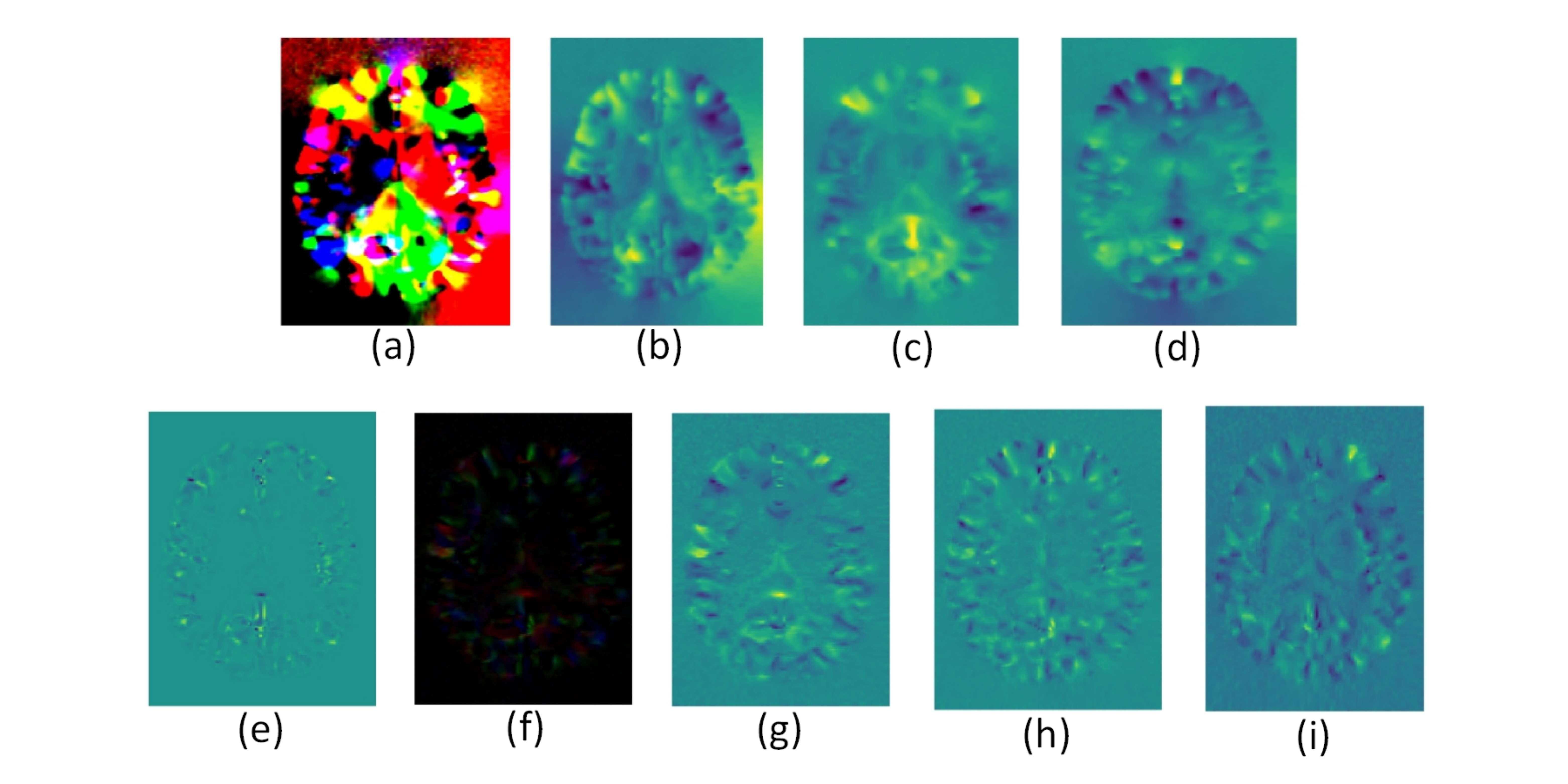}
\caption{Generated images based on the deformation field: (a)RGB image of deformation field $\phi$, (b)(c)(d)each channel of $\phi$, (e)JD image, (f) RGB image of CV image, (g)(h)(i) each channel of CV image.}
\end{figure}

\begin{figure}[ht]
\centering
\includegraphics[scale=1,width=0.5\textwidth]{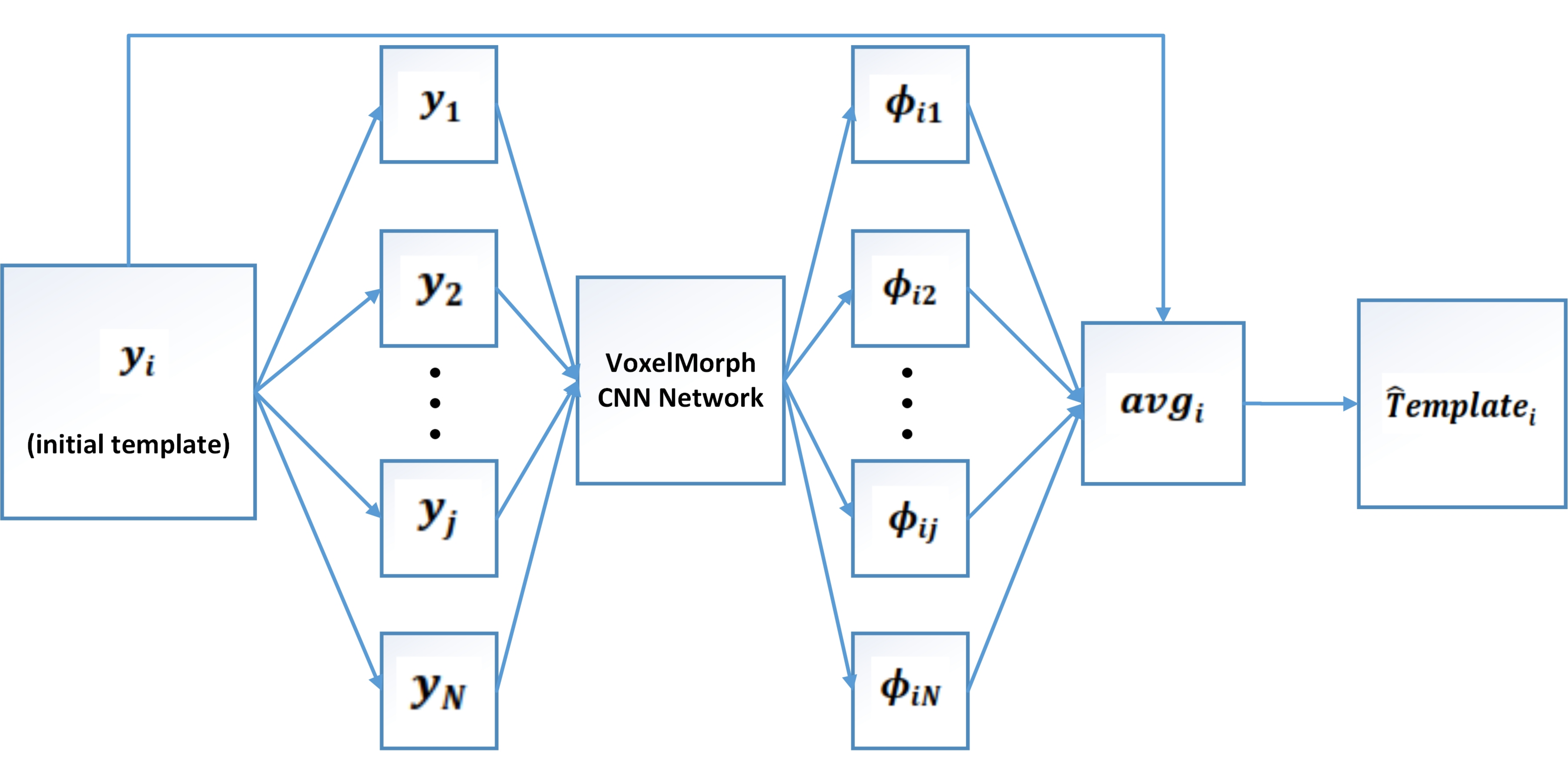}
\caption{Average of transformation method.}
\end{figure}

\subsubsection{Evaluation Criteria}
In this paper, Dice Score(DS) is the most used criteria for volume overlap of anatomical segmentations. We expect the regions in $M(\phi)$ and $F$ according to the same anatomical structure to overlap well. Let ${\mathrm{S}}^{k}_{F}$,${\mathrm{S}}^{k}_{M(\phi)}$ be the voxels for $M(\phi)$ and F respectively.DS is computed by:
\begin{equation}
DS({\mathrm{S}}^{k}_{F},{\mathrm{S}}^{k}_{M(\phi)})=2*{\frac{{\mathrm{S}}^{k}_{F}\cap{\mathrm{S}}^{k}_{M(\phi)}}{|{\mathrm{S}}^{k}_{F}|+|{\mathrm{S}}^{k}_{M(\phi)}|}}
\end{equation}
Dice score is between 0 and 1, and Dice score is closer to 1 indicates that the structures are more identical. And vice, a score of 0 indicates that there is no overlap. We also evaluate the regularity of the registration field $\phi$. The Jacobian matrix ${\mathrm{J}_{\phi}}(p)=\bigtriangledown\phi(p)\in{\mathbb{R}}^{3x3}$ captures the local properties of $\phi$.We compute the number of all non-background voxels for which $|{\mathrm{J}_{\phi}}(p)|\le0$, where $\phi$ is not diffeomorphic.
\subsubsection{Implementation}
Our implementation uses Keras[18] with a Tesorflow backend[12] and the Adam optimizer[14] with a learning rate of $1{\mathrm{e}}^{-4}$. We used MATLAB to generate the images formed by JD and CV information based on brain MRI, and saved it as nii image format. We set the epochs as 1500, batchsize as 1, steps of per epoch as 100 using one GeForce GTX 1080 Ti GPU. Each training batch consists of one pair of volumes to reduce the memory usage. We choose the model that optimizes Dice on validation set and get results on test set. Algorithm runtime were computed for a GeForce GTX 1080 Ti GPU and an Intel Core i7-6800k CPU.

\section{Results}
\subsection{Experiments on Test Set}
We present the following experiments demonstrating the results of our proposed method compared with state-of-the-arts methods. Specifically, we register each scan to an atlas computed using external data [6] which is also used on MIT's paper. Table 1 gives a summary of the results on test sets. According to Fig.2, we get JD and CV from two different $\phi$, one $\phi$ is directly generated using deformation method on the gradient and gray value of input moving volume described in section 2.1 and another is the registration filed $\phi$ in the first training stage, which we refer to two methods as VoxelMorph-deformation and VoxelMorph-registration. We compare them with state-of-the-arts methods the original VoxelMorph and VoxelMorph-diff from MIT's work [15][4]. The first method VoxelMorph-deformation get the best average Dice with 0.764. The latter three method have comparable runtimes but all faster and yield uncertainty estimates than original VoxelMorph. However, our second method, VoxelMorph-registration produces better diffeomorphic registration fields(having lower numbers of non-negative Jacobian locations) than other methods.
\begin{table}[H]
	\begin{center}
		\begin{tabular}{|*{6}{c|}}
			\toprule
			\multicolumn{1}{|c|}{Method} & \multicolumn{1}{|c|}{Avg. Dice} & \multicolumn{1}{|c|}{GPU sec} & \multicolumn{1}{|c|}{CPU sec} &
			\multicolumn{1}{|c|}{Percentage($\%$) of $|{\mathrm{J}_{\phi}}(p)|\le0$} &
			\multicolumn{1}{|c|}{Uncertainty}
			\\
			
			\midrule
			VoxelMorph(CC)[15] & 0.752(0.142) & 0.53(0.01) & 85.1(1.21) & 0.562(0.112) & No\\
			
			
			VoxelMorph-diff[4] & 0.753(0.138) & 0.45(0.01) & 56(0.13) & 0.511(0.182) & Yes\\
			
			VoxelMorph-deformation & 0.764(0.125) & 0.46(0.02) & 57(0.12) & 0.504(0.186) & Yes\\
			VoxelMorph-registration & 0.758(0.134) & 0.48(0.14) & 59(0.24) & 0.492(0.235) & Yes\\
			
			VoxelMorph-atlas & 0.756(0.141)& 0.41(0.23) & 58(0.22) & 0.487(0.194) & Yes\\

			\bottomrule
		\end{tabular}
	\end{center}
	
	\caption{Test average results of IBSR training set(10-14) for different N of JD images with three modalities(DC:\%, HD: mm, AVD:\%).}
\end{table}

Fig.5 shows representative results. We present the example MR slices (Two anatomical planes) of input
moving image, atlas, and moved images for different experiments, with overlaid boundaries of ventricles (yellow). From the red box, according to atlas, our latter two methods get clearer details of moved images and better registration performance than VoxelMorph-diff. Fig.6 shows the registration fild $\phi$ for different experiments (RGB image and warped grid), the method VoxelMorph-registration get smoother diffeomorphic registration field $\phi$ than else two method.
\begin{figure}[ht]
\centering
\includegraphics[scale=1,width=0.5\textwidth]{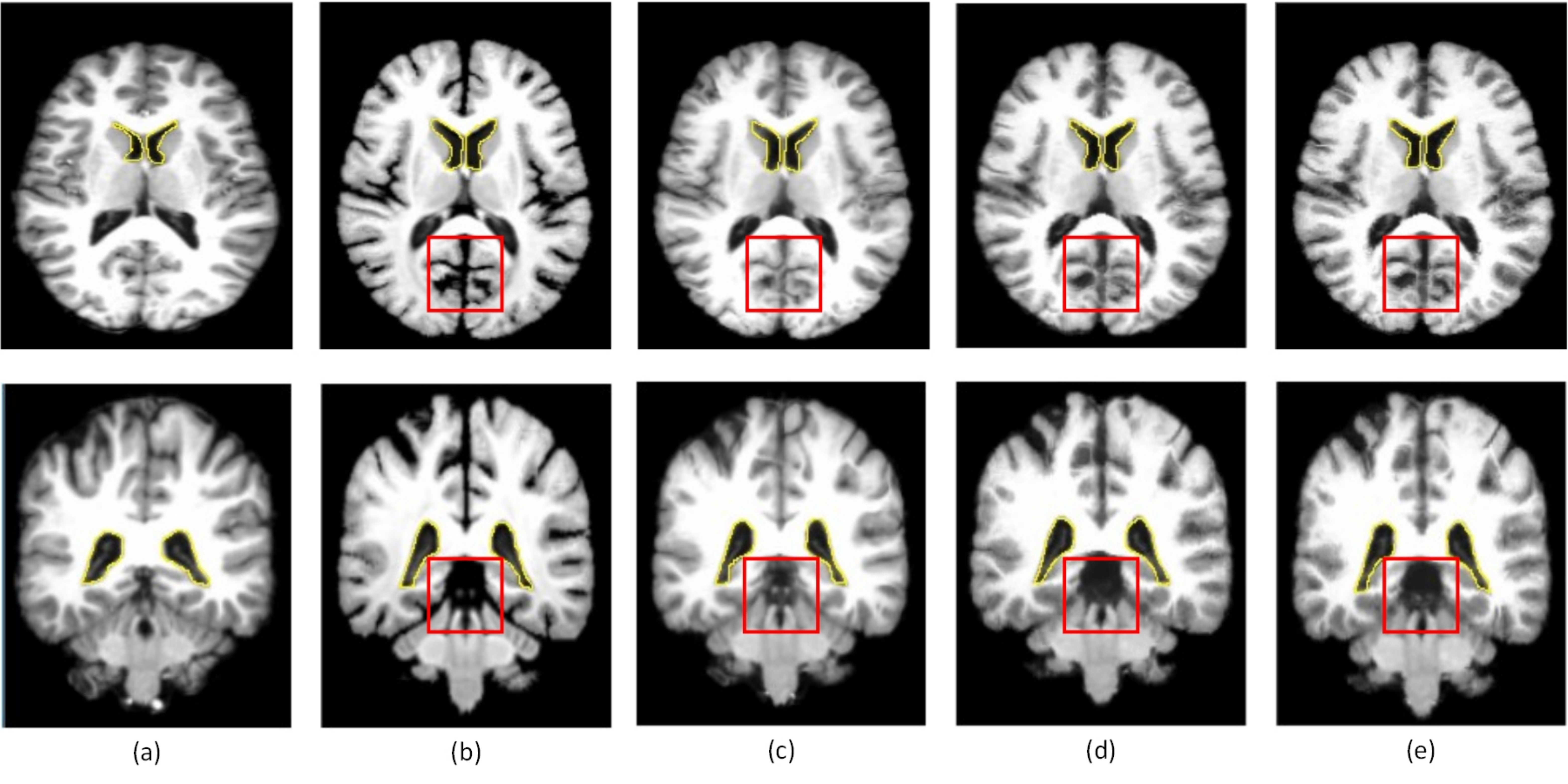}
\caption{Comparison of example MR slices (Two anatomical planes) for different experiments: (a)moving image, (b)atlas (fixed image), (c) moved (VoxelMorph-diff), (d) moved (VoxelMorph-registration), (e) moved (VoxelMorph-deformation).}
\end{figure}

Based on FreeSurfer segmentation, we used three methods already trained to test the anatomical scans from the unseen anatomical structures of ADNI. This dataset contains expert manual delineations of anatomical structures, which is used for our evaluation. The following experiments shows the results of anatomical segmentation for different experiments. We can see VoxelMorph-deformation method performs better segmentation results than other methods from the different color regions.
\begin{figure}[ht]
	\centering
	\includegraphics[scale=1,width=0.5\textwidth]{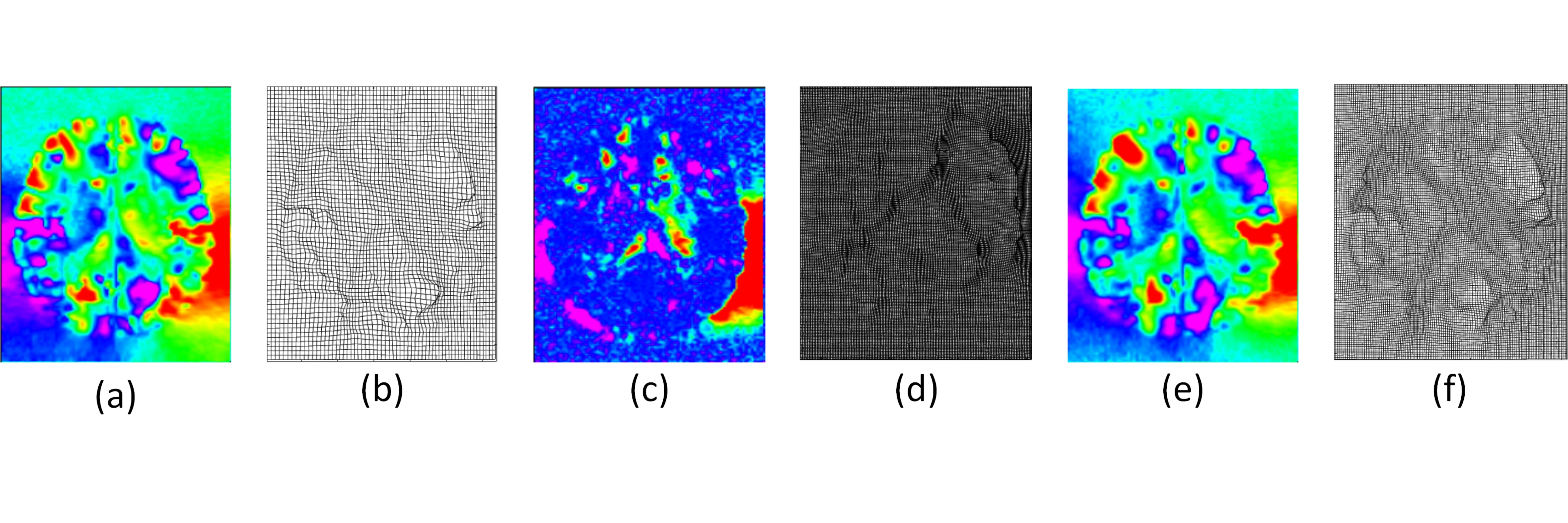}
	\caption{Comparison of registration filed $\phi$ for different experiments(RGB image and warped grid): (a)(b) $\phi$ from VoxelMorph-diff, (c)(d) $\phi$ from VoxelMorph-registration, (e)(f)$\phi$ from VoxelMorph- deformation.}
\end{figure}

\begin{figure}[ht]
	\centering
	\includegraphics[scale=1,width=0.5\textwidth]{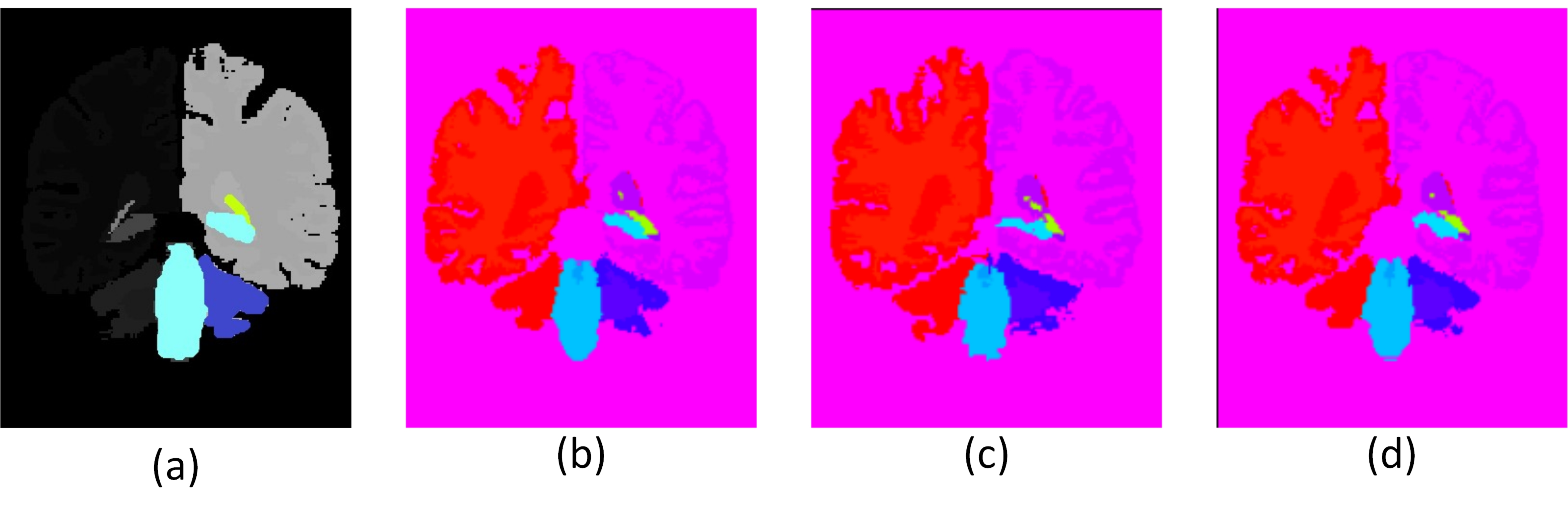}
	\caption{Results of anatomical segmentation for different experiments: (a)anatomical labels from fixed image, (b) segmentation from VoxelMorph-diff, (c) segmentation from VoxelMorph-registration, (d) segmentation from VoxelMorph- deformation.}
\end{figure}
\begin{figure}[ht]
	\centering
	\includegraphics[scale=1,width=0.5\textwidth]{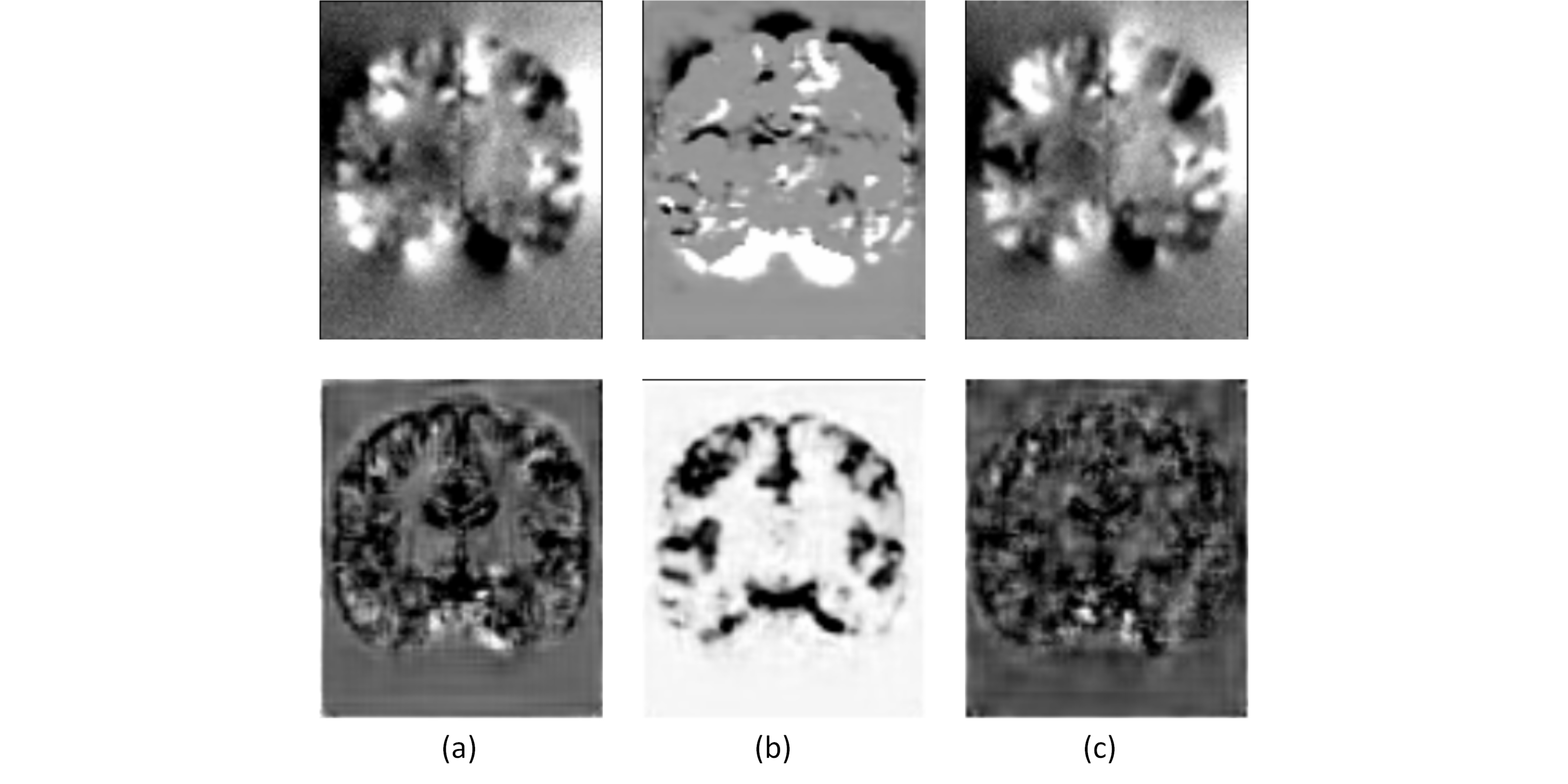}
	\caption{Analysis of velocity sampling and uncertainty for different experiments(top is velocity sample ${\mathrm{z}}_{k}$ and bottom is variational covariance ${\varSigma}_{z|x,y}$: (a) VoxelMorph-diff, (b) VoxelMorph-registration, (c) VoxelMorph- deformation.}
\end{figure}
\subsection{Analysis of Velocity Sampling and Uncertainty}
In this section, we evaluate generated velocity sampling and uncertainty for different experiments. From the bottom row, we can see the VoxelMorph-registration method learns smaller values for ${\varSigma}_{z|x,y}$ approximation and yields smoother velocity sample ${\mathrm{z}}_{k}$ than other two methods, namely the diagonal covariance ${\varSigma}_{z|x,y}$ has the less potential to add noise to ${\mathrm{z}}_{k}$, so, the loss function coupled with the squaring and scaling layers lead to smoother and more accurate registration field $\phi$ as Fig.6 shows. And this also testify the result that it can generate better diffeomorphic registration fields which shown in Table 1.

\subsection{Experiments on Atlas}
We test the average transformation method using MRBrainS18 dataset to construct a template(atlas) for
7 samples and use it as fixed images for medical registration based on above original VoxelMorph-diff
network. We do average transformation method for two iterations, the two average registration field are shown in Fig.9 (a)(b) respectively. Because the average transformation ${\mathrm{avg}}_{2}$ is close to the identify map Id, so we stop the iteration step and use one of the Template as fixed image for original VoxelMorph-diff network. From the Table 1, we refer this experiment as VoxelMorph-atlas, compared with VoxelMorph-diff, it has comparable average Dice scores, runtimes and yield uncertainty estimates but produces better diffeomorphic registration fields(having lower numbers of non-negative Jacobian locations, $0.487(0.194)\%$ for VoxelMorph-atlas and $0.511(0.182)\%$ for VoxelMorph-diff). Moving, fixed image(atlas), fixed image and generated registration field $\phi$ are shown in Fig.9(c)(d)(e)(f) respectively.

\begin{figure}[ht]
\centering
\includegraphics[scale=1,width=0.5\textwidth]{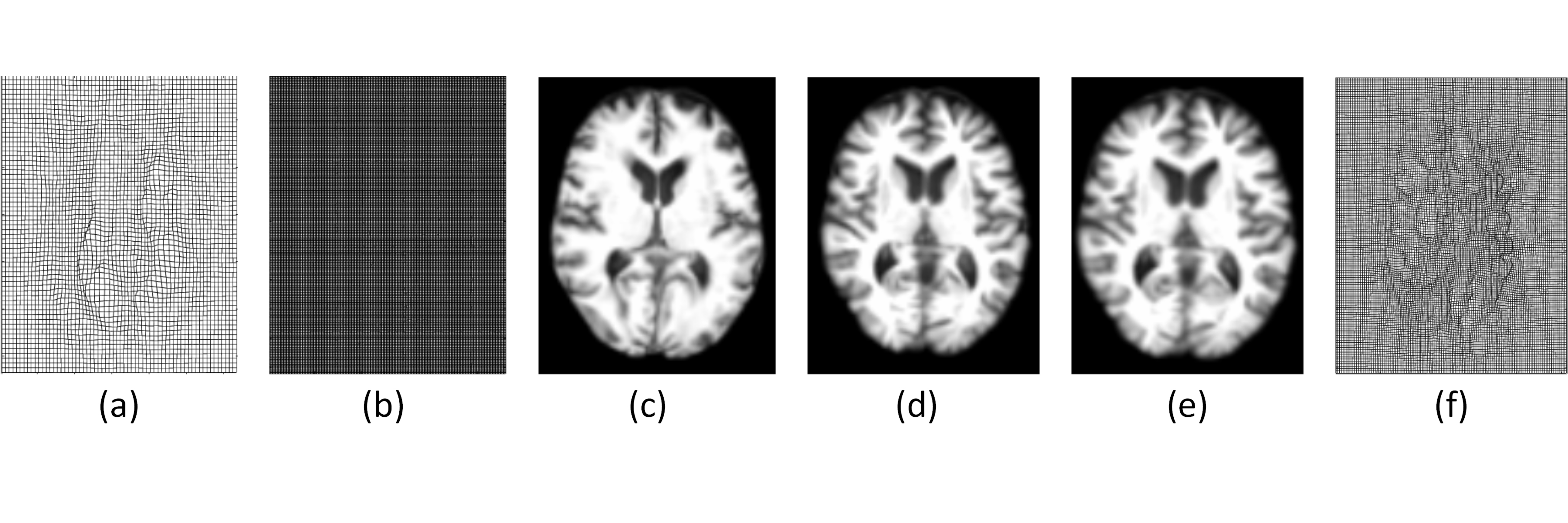}
\caption{Results of atlas construction: (a)registration field after first iteration, (b) registration field after second iteration, (c)moving image, (d) constructed atlas as fixed image, (e)moved image, (f) resulting registration field.}
\end{figure}

\section{Conclusions}
In conclusion, we propose a new deformable medical image registration method based on differential geometry and VoxelMorph CNN architecture. We compute the differential geometric information including the Jacobian determinant (JD) and the curl vector(CV) of diffeomorphic registration field and use them as channels of the model to train again. We get JD and CV from two sources, generated diffeomorphisms $\phi$ directly by original MR images and the registration filed $\phi$ in the first training stage, we test our method on two datasets including ADNI dataset and the results show our method VoxelMorph-deformation has better Dice and VoxelMorph-registration has better diffeomorphic registration fields than MIT's result. However, our method needs amounts of iterations and mathematical computation when generating JD and CV.

We also build a registration template for 7 testing samples according to average transformation method based on MRBrainS18 Challenge dataset, and obtain excellent improvement on medical image registration with non-negative Jacobian locations compared with MIT's VoxelMorph-diff. However, it also needs some iteration and computing resource which depends on the volume numbers of dataset. It may be useful to construct an enormous atlas dataset for medical registration. We believe the proposed method can advance the performance in medical image registration and clinical diagnosis.



\begin{thebibliography}{1}
	\providecommand{\url}[1]{#1}
	\csname url@samestyle\endcsname
	\providecommand{\newblock}{\relax}
	\providecommand{\bibinfo}[2]{#2}
	\providecommand{\BIBentrySTDinterwordspacing}{\spaceskip=0pt\relax}
	\providecommand{\BIBentryALTinterwordstretchfactor}{4}
	\providecommand{\BIBentryALTinterwordspacing}{\spaceskip=\fontdimen2\font plus
		\BIBentryALTinterwordstretchfactor\fontdimen3\font minus
		\fontdimen4\font\relax}
	\providecommand{\BIBforeignlanguage}[2]{{%
			\expandafter\ifx\csname l@#1\endcsname\relax
			\typeout{** WARNING: IEEEtranS.bst: No hyphenation pattern has been}%
			\typeout{** loaded for the language `#1'. Using the pattern for}%
			\typeout{** the default language instead.}%
			\else
			\language=\csname l@#1\endcsname
			\fi
			#2}}
	\providecommand{\BIBdecl}{\relax}
	\BIBdecl
	
	\bibitem{[1]}
	~Bajcsy R. \& ~Kovacic, S. Multiresolution elastic matching. \emph{Computer Vision, Graphics, and Image Processing}.\textbf{46},1--21(1989).
	
	\bibitem{[2]}
	~Zhu Y.P., ~Zhou Z.C., ~Liao G.J., ~Yang Q.X., ~Yuan K.H.
	Effects of Differential Geometry Parameters on Grid Generation and Segmentation of MRI Brain Image. \emph{IEEE Access}.\textbf{7}(1),68529--68539(2019)
	
	\bibitem{[3]}
	~Dalca A.V., ~Balakrishnan G., ~Guttag J., ~Sabuncu M.R. Unsupervised Learning for Fast Probabilistic Diffeomorphic Registration. arXiv preprint arXiv: 1805.04605, 2018.
	
	\bibitem{[4]}
	~Dalca A.V., ~Balakrishnan G., ~Guttag J., ~Sabuncu M.R. Unsupervised Learning of Probabilistic Di?eomorphic Registration for Images and Surfaces. arXiv preprint arXiv: 1903.03545, 2019.
	
	\bibitem{[5]}
	~Adrian V.D., ~Andreea B., ~Natalia S.R., ~Polina G. Patch-based discrete registration of clinical brain images.In \emph{MICCAI-PATCHMI Patch-based Techniques in Medical Imaging.Springer}(2016).
	
	\bibitem{[6]}
	~Fischl B. Freesurfer. \emph{Neuroimage}.\textbf{62}(2),774--781(2012)
	
	\bibitem{[7]}
	~Daniel R., ~Luke I.S., ~Carmel H., ~Derek L.G.H., ~Martin O.L., ~David J. H. Nonrigid registration using free-form deformation: Application to breast mr images.\emph{IEEE Transactions on Medical Imaging}.\textbf{18}(8),712--721(1999).
	
	\bibitem{[8]}
	~Susanne G.M., ~Michael W.W., ~Leon J.T., ~Ronald C.P., ~Cli?ord R.J., ~William J., ~John Q.T., ~Arthur W.T., ~Laurel B. Ways toward an early diagnosis in Alzheimers disease: the Alzheimers Disease Neuroimaging Initiative (ADNI).\emph{Alzheimer?s \& Dementia}.\textbf{1}(1),55--66(2005).
	
	\bibitem{[9]}
	~Thirion J.P.Image matching as a di?usion process: an analogy with maxwell?s demons.\emph{Medical Image Analysis}.\textbf{2}(3),243?260(1998).
	
	\bibitem{[10]}
	~Lei N., ~Luo Z.X., ~Yau S.T., ~Gu X.F. Geometric Understanding of Deep Learning. arXiv preprint arXiv: 1805.10451,2018.
	
	\bibitem{[11]}
	~Balakrishnan G., ~Zhao A., ~Sabuncu M. R., ~Guttag J., ~Dalca A. V.
	An unsupervised learning model for deformable medical image registration.In \emph{Proceedings of the IEEE Conference on Computer Vision and Pattern Recognition}. pp. 9252--9260(2018)
	
	\bibitem{[12]}
	~Martin A., ~Ashish A., ~Paul B., ~Eugene Brevdo, ~Zhifeng Chen, ~Craig Citro, ~Greg S Corrado, ~Andy Davis, ~Je?rey Dean, ~Matthieu Devin, et al. Tensor?ow: Large-scale machine learning on heterogeneous distributed systems. arXiv preprint arXiv:1603.04467, 2016.
	
	\bibitem{[13]}
	~Chen, X., ~Liao, G. New variational method of grid generation with prescribed jacobian determinant and prescribed curl. \emph{Computer Science}.2--6(2015)
	
	\bibitem{[14]}
	~Diederik P.K., Jimmy B. ADAM: A method for stochastic optimization. arXiv preprint arXiv:1412.6980, 2014.
	
	\bibitem{[15]}
	~Balakrishnan G., ~Zhao A., ~Sabuncu M.R., ~Guttag J., ~Dalca A.V. VoxelMorph: A Learning Framework for Deformable Medical Image Registration. \emph{IEEE Transactions on Medical Imaging}.\textbf{99},1--1(2019)
	
	\bibitem{[16]}
	~Zhou Z.C., ~Hildebrandt B., ~Chen X., ~Liao GJ.Computational Technologies for Brain Morphometry.arXiv preprint arXiv: 1810.04833, 2018.
	
	\bibitem{[17]}
	~Ashburner J., ~Friston K.Voxel-based morphometry-the methods.\emph{Neuroimage}.\textbf{11},805--821(2000).
	
	\bibitem{[18]}
	~Francois C.Keras. https://github.com/fchollet/keras, 2015
	
	
	
\end{thebibliography}
%
\end{document}